\pdfoutput=1
\documentclass[11pt]{article}

\usepackage[final]{acl}

\usepackage{times}
\usepackage{latexsym}

\usepackage{xcolor}

\usepackage{amsmath}

\usepackage{booktabs, multirow} % for borders and merged ranges
\usepackage{soul}% for underlines
\usepackage{changepage,threeparttable} % for wide tables

\usepackage[T1]{fontenc}
\usepackage[utf8]{inputenc}

\usepackage{microtype}
\usepackage{amssymb}

\usepackage{inconsolata}

\usepackage{graphicx}

\usepackage{xtab}
\usepackage{tabularx} 
\usepackage{stfloats}
\usepackage[most]{tcolorbox}
\usepackage{float}
\usepackage{adjustbox}
\usepackage{makecell}
\usepackage{enumitem}
\usepackage{siunitx}

\definecolor{navyblue}{RGB}{0, 0, 128}
\definecolor{charcoal}{RGB}{54, 69, 79}
\definecolor{iceblue}{RGB}{240, 248, 255}
\definecolor{oxfordblue}{RGB}{4, 30, 66}
\definecolor{cream}{RGB}{255, 253, 208}
\definecolor{maroon}{RGB}{128, 0, 0}
\definecolor{forestgreen}{RGB}{34, 139, 34}
\definecolor{palecoral}{RGB}{240, 128, 128}
\definecolor{color1}{HTML}{00b4d8}
\definecolor{color2}{HTML}{90e0ef}
\definecolor{color3}{HTML}{caf0f8}

\title{Effective Self-Mining of In-Context Examples for Unsupervised Machine Translation with LLMs}

\author{\normalsize Abdellah {El Mekki} $^{\xi}$ ~~~~~~~~ Muhammad Abdul-Mageed $^{\lambda, \xi,\gamma}$\\
$^{\lambda}$The University of British Columbia ~~~~~~~~~ \normalsize $^{\xi}$MBZUAI ~~~~~~~~  $^{\gamma}$Invertible AI \\
\texttt{abdellah.elmekki@mbzuai.ac.ae, muhammad.mageed@ubc.ca}
} 

\begin{document}
\maketitle
\begin{abstract}
Large Language Models (LLMs) have demonstrated impressive performance on a wide range of natural language processing (NLP) tasks, notably through in-context learning (ICL). In ICL, an LLM is provided with examples that represent a given task such that it learns to generate answers for test inputs. However, access to these in-context examples is not guaranteed especially for low-resource or massively multilingual tasks. In this work, we propose an unsupervised approach to mine in-context examples for machine translation (MT), enabling unsupervised MT (UMT) across different languages.
Our approach begins with word-level mining to acquire word translations that are then used to perform sentence-level mining. As the quality of mined parallel pairs may not be optimal due to noise or mistakes, we introduce a filtering criterion to select the optimal in-context examples from a pool of unsupervised parallel sentences.
We evaluate our approach using two multilingual LLMs on 288 directions from the FLORES-200 dataset~\cite{nllbteam2022language} and analyze the impact of various linguistic features on performance. Our findings demonstrate the effectiveness of our unsupervised approach in mining in-context examples for MT, leading to better or comparable translation performance as translation with regular in-context samples (extracted from human-annotated data), while also outperforming the other state-of-the-art UMT methods by an average of $7$ BLEU points. Our code is available at: \href{https://github.com/UBC-NLP/sm-umt}{https://github.com/UBC-NLP/sm-umt}. %, particularly in language pairs where it is hard to have access to parallel examples.
\end{abstract}

\section{Introduction}

Large language models (LLMs) have significantly advanced various natural language processing (NLP) tasks \cite{10.5555/3495724.3495883, DBLP:journals/corr/abs-2001-08361, ouyang2022training, touvron2023llama}, with generative pretrained Transformer (GPT) models, which follow a decoder-only architecture~\cite{Radford2018ImprovingLU, radford2019language}, demonstrating particularly admirable outcomes. Performance of LLMs is especially effective when using few-shot learning through \textit{in-context learning} (ICL, \citealt{10.5555/3495724.3495883}). In ICL, the LLM is provided with task-specific input-output examples, allowing it to learn how to make predictions for a test input based on these examples without further fine-tuning. This approach has shown promising results in tasks such as question answering~\cite{li-etal-2023-shot}, common sense reasoning~\cite{10.1162/tacl_a_00370, wu-etal-2023-self}, and text classification~\cite{milios-etal-2023-context, khondaker2023gptaraeval, chandra2024one}. However, ICL is sensitive to several constraints. For example, it is highly dependent on the selected in-context samples. In addition, small changes in the prompt can lead to high variance in acquired outputs~\cite{lu-etal-2022-fantastically, chang-jia-2023-data}.
To address these issues, researchers have proposed several methods to choose the best examples that can lead to optimal performance~\cite{rubin-etal-2022-learning, li-etal-2023-unified, luo2023dricl, agrawal-etal-2023-context}. For instance,~\citet{agrawal-etal-2023-context} propose selecting examples that are closer to the test input in terms of BM25 score. Another challenge is that performance of ICL can be affected if the task is expressed in a language not well represented in the LLM pretraining data~\cite{huang-etal-2023-languages}. This is especially relevant to work attempting to leverage multilingual LLMs such as XGLM \cite{lin-etal-2022-shot}, Bloom \cite{workshop2023bloom}, Gemma \cite{gemmateam2024gemma}, and Llama-3~\cite{llama3modelcard}. %, finding suitable in-context examples for each language remains challenging for different tasks.
% In this paper, we focus on LLMs for the machine translation (MT) task, where we assume that \underline{in-context examples are not available}; a scenario that often occurs when a low-resource language is either the source or target. 

While previous efforts have demonstrated the effectiveness of in-context learning through careful example selection, this approach assumes that relevant in-context examples are readily available. However, in tasks such as machine translation (MT), especially for low-resource languages, this assumption often does not hold, leading to challenges in leveraging LLMs effectively. In this paper, we address this limitation by focusing on \textit{unsupervised machine translation} (UMT) in scenarios where in-context examples are not available. Recent studies have tackled UMT by leveraging monolingual data available in the source and target languages~\cite{lample2018word, artetxe-etal-2019-effective} using techniques such as alignment of word embeddings~\cite{lample2018word} and back-translation~\cite{sennrich-etal-2016-improving, lu-zhang-2024-improving}. In contrast, our approach assumes access to only a small portion of unlabeled data in each language (less than 1,000 sentences) and a multilingual LLM. To tackle the UMT problem, we divide it into two stages: (i) \textbf{a word-level translation stage}, where we use the LLM to generate high-quality word translations that we exploit to create synthetic word-by-word translated parallel data; and (ii) \textbf{a sentence-level translation stage}, where we leverage the synthetic parallel data to create better in-context examples that we can use to translate the test input. Additionally, we study several methods for in-context example selection from the unsupervised mined sentence pairs.

We experiment with two multilingual LLMs as our base testbeds, namely, Llama-3 (8B)~\cite{llama3modelcard} and Bloom (7B)~\cite{workshop2023bloom}. We conduct extensive experiments on the FLORES-200~\cite{nllbteam2022language} evaluation sets, considering three major differences between languages: language \textit{resource level}, language \textit{family}, and language \textit{script}. To summarize, we offer the following contributions:

\begin{itemize}
    \item We propose a two-stage unsupervised method to extract in-context examples for any language pair. Initially, our approach involves word-level translation, which is subsequently expanded to sentence-level translation.
    
    \item We propose an input-specific in-context example selection method for unsupervised mined sentence pairs, which combines sentence embedding similarity scores and BM25~\cite{10.1145/2682862.2682863} to identify the most relevant and informative examples for each test input sentence.
    
    \item We perform extensive evaluations on $288$ language directions covering various \textit{resource levels} (High, Medium, Low, and Very Low), \textit{scripts} (e.g., Latin, Arabic, Hangul), and \textit{language families} (e.g., Romance, Afro-Asiatic, Turkic) of the source and target languages.

    \item We compare our method to robust baselines such as existing state-of-the-art UMT methods and regular (and BM25) $k$-shot ICL MT. Our unsupervised method not only outperforms previous UMT approaches but also achieves results comparable to, or surpassing, those of the regular ICL method in most of the experiments we conduct.

    \item We conduct an extensive analysis to explore the factors influencing UMT performance, focusing on resource availability, language family, and script. Additionally, we examine various linguistic features, including language distances, uncovering notable correlations between these features and UMT performance.
\end{itemize}

\section{Related work}

\paragraph{Unsupervised Word Translation.} Bilingual lexicon induction (BLI) is a crucial task for UMT \cite{artetxe-etal-2018-unsupervised, marchisio-etal-2020-unsupervised}. It is the task of inducing word translations across languages from monolingual corpora. BLI approaches can be categorized as either supervised or unsupervised. Supervised methods rely on a seed dictionary, monolingual word embeddings, and an alignment algorithm to learn the alignment between the source and target languages \cite{artetxe-etal-2016-learning, 10.5555/3504035.3504649}. In contrast, unsupervised methods rely solely on monolingual word embeddings and an unsupervised alignment algorithm \cite{lample2018word, artetxe-etal-2018-robust}.
Recent work has started exploring the use of multilingual LLMs in BLI by prompting or fine-tuning them to translate words from one language to another \cite{li-etal-2023-bilingual, el-mekki-etal-2023-promap, li2024selfaugmented}, leading to results better than relying on static word embeddings alone. More recently, \citet{li2024selfaugmented} proposed a fully unsupervised method that leverages an in-context seed dictionary to translate words using multilingual LLMs. %Their method has been shown to outperform all previous methods, including supervised ones.

\paragraph{Unsupervised Machine Translation (UMT).} %is the task of translating between two languages without the need for parallel sentences.
A key technique that has gained prominence in recent UMT research is back-translation, first introduced by \citet{sennrich-etal-2016-improving}. This approach involves translating monolingual data in the target language to create a synthetic parallel corpus for training, which is especially beneficial for language pairs with limited parallel examples.
Modern UMT approaches have made remarkable progress by incorporating representation learning \cite{lample2018word} and pre-trained generative models \cite{artetxe-etal-2018-unsupervised, song2019mass, he-etal-2022-bridging,lu-zhang-2024-improving} into the existing framework of BLI and back-translation. These advancements allow UMT models to learn cross-lingual representations and generate translations without the need for parallel supervision.
%UMT models employ iterative learning to continuously refine translation quality \cite{hoang-etal-2018-iterative, Guo2020RevisitingIB}.
Recently, a study by \citet{10.5555/3618408.3620130} has introduced a novel perspective on UMT by showcasing the potential of LLMs in creating pseudo-parallel examples through prompting leading to better UMT performance.

\paragraph{In-Context Examples Selection.} Several methods have been proposed to select optimal in-context examples for language tasks, including MT \cite{liu-etal-2022-makes, rubin-etal-2022-learning, li-etal-2023-unified, luo2023dricl}, often relying on BM25 or fine-tuned dense retrievers to find examples similar to the test input. In MT, \citet{agrawal-etal-2023-context} and \citet{sia-duh-2023-context} found that selecting examples with greater n-gram overlap or from the same domain improves translation performance, with longer examples generally yielding better results. However, \citet{10.5555/3618408.3620130} showed that MT performance does not always correlate with the selected in-context examples, noting high variance even with the same examples, and that $1$-shot learning can underperform zero-shot approaches.
To address example selection challenges, \citet{m-etal-2023-ctqscorer} proposed a scoring model evaluating sentence pairs based on semantic similarity, translation quality, and perplexity to select top-K in-context examples, but it requires annotated data and tools trained on parallel data. \citet{nguyen-etal-2024-democratizing} introduced Linguistically-Diverse Prompting (LDP), leveraging LLMs and synthetic exemplars from high-resource languages. For X→English translation, it uses diverse language pairs; for other directions, a two-step back-translation process. They found that using diverse languages for exemplars is more effective than relying on a single related language and helps models identify the correct target language.
In contrast, we focus on unsupervised mining of parallel pairs, proposing a method to filter and select optimal examples from generated synthetic data to enhance UMT performance using LLMs.

%\section{Background \textcolor{red}{(To be reformulated)}} 

%\paragraph{ICL Prompting.} Generating text with LLMs like LLama, Bloom, and Gemma for a specific NLP task involves conditioning the decoder-only language model by using in-context examples, this enables them to perform NLP tasks without any parameter updates or fine-tuning. These in-context examples fulfill two functions: (i) they inform the model about the structure and nature of the task at the task level, and (ii) they steer the generation of outputs by supplying pertinent details regarding the unseen source sentence, which is specific to the example.

%In simple terms, if we have \(k\) examples \(\{(x_i, y_i)\}_{i=1}^k\), we can concatenate them to build the prompt, \(x_p^{(j)}\), that we feed to the decoder-model. This is done by adding the example pairs \(\{(x_i, y_i)\}_{i=1}^k\) to a new test input, \(x_s^{(j)}\), following a specific format. The result, \(\hat{y}_j\), is then created using a predictive model with settings \(\theta\) and by choosing the most likely option at each step.

\section{Method} \label{sec:method}

\paragraph{UMT Task Description.} We consider two distinct languages, denoted as $L_s$ (source) and $L_t$ (target), each with access to its corresponding most frequent words vocabulary $V_s$ and $V_t$, respectively. Additionally, we assume access to a set of unlabeled sentences $D_U$ in $L_t$. Access to a multilingual LLM is also assumed. Our UMT method uses these resources to develop an MT system that translates sentences from $L_s$ to $L_t$ without needing human-labeled parallel data. Furthermore, we assume access to a test set $D_T$, containing parallel sentences between $L_s$ and $L_t$, which we use for evaluation. Importantly, \textit{no parallel data are available for the learning phase}. Meanwhile, we also assume access to an unsupervised similarity function $sim(x,y)$ that computes the cosine similarity between sentences $x$ and $y$, each from a different language. In this work, we focus on UMT using ICL of a multilingual LLM. Our task involves the unsupervised extraction (mining) of $k$ examples, which will be utilized as in-context examples to perform MT between $L_s$ and $L_t$, as getting these examples is still a challenge for under-represented languages. We now describe our method.

\paragraph{Methodology.} Our primary objective is to mine an effective set of $k$ parallel sentences in an unsupervised manner. These $k$ examples can then be used to translate input sentences from the source language $L_s$ in the test set $D_T$. Our approach consists of two main phases: (i) a \textit{word-by-word} translation phase, and (ii) a \textit{sentence-level} translation phase. It begins with performing word-level translation that we can leverage to produce synthetically annotated parallel data, which we use to initialize our sentence-level MT system.

\subsection{Word-by-Word Translation Phase}
A typical word translation task requires access to a seed dictionary containing word pairs between $L_s$ and $L_t$. Then, one can use a word embedding mapping algorithm~\cite{artetxe-etal-2016-learning, 10.5555/3504035.3504649} or fine-tune a language model~\cite{el-mekki-etal-2023-promap, li-etal-2023-bilingual} to learn the word translation task. However, in the absence of such a dictionary, we rely on LLMs to achieve this. In this section, we describe the process we follow to build our word translation model in an unsupervised manner.

\subsubsection{Word Pair Mining}\label{wpm}
We adopt an approach inspired by~\citet{li2024selfaugmented} to perform unsupervised word translation using LLMs. Given a multilingual LLM, we can mine word pairs from it using zero-shot prompting. To achieve this, we take each word from the source vocabulary $V_s$ and prompt the LLM to translate it from $L_s$ to $L_t$ using the following prompt template:\footnote{We experimented with several prompt templates and found that the presented one yields optimal results. The templates are inspired by \citet{10.5555/3618408.3620130} and \citet{li2024selfaugmented}.}

\begin{quote}
    ``The \( L_s \) word "\( w_s \)" in \( L_t \) is:'', 
\end{quote}

\noindent where $L_s$, $L_t$, and $w_s$ are placeholders for the \textit{source} language, \textit{target} language, and the \textit{query} word, respectively.

Following a random sampling decoding (we discuss the use of other decoding strategies in Figure \ref{fig:duration_decoding}, Appendix \ref{sec:app_anal}), the LLM can generate $n$ sequences (generation stops upon reaching the space token)\footnote{We follow this method as it aligns with established static word embedding alignment methods for BLI, which typically focus on single-word alignments.} ranked by their sequence scores. Then, we filter out the predicted translations that are not included in the vocabulary $V_t$. This step generates a pool of word pairs $P_{s \rightarrow t}$, containing a maximum of $n \times |V_s|$ pairs (as some pairs are filtered out). The next step is to reverse the process, going from the target language back to the source language (using greedy sampling decoding), focusing on target words in $P_{s \rightarrow t}$. This results in another pool of word pairs, $P_{t \rightarrow s}$, from $L_t$ to $L_s$.

Finally, we identify and retain the best pairs from both $P_{s \rightarrow t}$ and $P_{t \rightarrow s}$ that accurately represent the word translations between the source language $L_s$ and the target language $L_t$. Unlike \citet{li2024selfaugmented}, this selection is performed in two steps:

\begin{enumerate}[noitemsep,topsep=0pt]
    %\item The first step involves retaining only the pairs $(w_s, \hat{w}_t)$ that exist in both $P_s$ and $P_t$. Specifically, we keep the pair $(w_s, \hat{w}_t)$ if and only if $w_s$ results from the back-translation of $\hat{w}_t$.% from $L_t$ to $L_s$.
    
    %\item The second step involves ranking the pairs according to their cosine similarity scores, computed using the function $sim(w_{s},\hat{w}_t)$. The top $k$ pairs are selected. The objective of this step is to sort the pairs based on their semantic similarity, thereby ensuring a higher quality of word pairs.
    \item Keep pairs $(w_s, \hat{w}_t)$ present in both $P_{s \rightarrow t}$ and $P_{t \rightarrow s}$, where $w_s$ is the back-translation of $\hat{w}_t$.
    
    \item Rank pairs by cosine similarity scores using $sim(w_{s},\hat{w}_t)$. Select top $k_{wp}$ pairs. This sorts pairs by semantic similarity, ensuring a higher quality of word pairs.

\end{enumerate}

\subsubsection{$k$-Shot Word Pair Mining}

We replicate the steps from the previous mining process (i.e., in Section~\ref{wpm}), but with two key modifications. First, we substitute zero-shot prompting with $k$-shot ICL. Second, we employ the top $k_{wp}$ word pairs selected in Section~\ref{wpm} as examples for the ICL process, essentially bootstrapping the learning from our initial results. This step produces a more refined version of $k_{wp}$ word pairs, leveraging the knowledge gained from the initial selection to improve our word-by-word sentence translation accuracy and reliability in subsequent steps.

\iffalse
The new prompt template is the following:
%,title=\textbf{$k$-Shot Word Translation Prompt Template}
\begin{tcolorbox}[colback=iceblue,colframe=oxfordblue]
\textcolor{charcoal}{\begin{quote}
Translate the following \( L_s \) word to \( L_t \): \\
\hspace*{2em} \textbf{\(L_s\)}: \(w_{s_1}\) \\
\hspace*{2em} \(L_t\): \(w_{t_1}\) \\
\hspace*{2em} \color{black}........ \\\color{charcoal}
\hspace*{2em} \(L_s\): \(w_{s_k}\) \\
\hspace*{2em} \(L_t\): \(w_{t_k}\) \\
\hspace*{2em} \(L_s\): \(w_{s_t}\) \\
\hspace*{2em} \(L_t\): \\,
\end{quote}}
\end{tcolorbox}

\noindent where $L_s$, $L_t$, \(\{(w_{s_i}, w_{t_i})\}_{i=1}^k\), and $w_{s_t}$ are placeholders for the source language, target language, $k$ word pairs examples, and the query word, respectively.
\fi

%To further enhance the quality of the mined word pairs, this step can be iterated multiple times \footnote{\textcolor{red}{In our experiments, two iterations typically yielded optimal results, balancing efficiency and performance gains. The marginal improvement obtained from a second iteration was generally insufficient to justify additional iterations' computational cost. Therefore, we conducted a single iteration for the reported experimental results.}}.

% we iterate this $k$-shot mining process multiple times. Each iteration leverages the most relevant and high-quality word pairs from the previous round, serving as an updated set of examples and enabling the mining of even higher-quality pairs in the next iteration.

\subsubsection{Weakly-Annotated Synthetic Parallel Data.} \label{sec:w2w}
We use the $k_{wp}$ mined word pair as in-context examples to translate words in each sentence of $D_T$, resulting in a set of sentences $D_{w2w}$ that have been translated word-by-word from $L_s$ to $L_t$. Although these translations do not maintain accurate word order or grammatical structure, we hypothesize they would preserve the semantic meaning of the original sentences.

\subsection{Sentence-Level Translation Phase} \label{sec:sent_trans}
In the following, we use $D_{w2w}$ to generate more accurate translations. The process is as follows:

\subsubsection{$L_t$ to $L_s$ Back-Translation}\label{sec:tran_step_1}

Given that the sentences in $D_{w2w}$ are translated on a word-by-word basis, the quality of these translations may be suboptimal compared to the ground truth, with potential issues such as missing words, incorrect word order, or simple inaccuracies in the translation. However, the overall meaning is observably largely preserved. To address these issues, we follow a back-translation step: we reverse the translation direction by using the word-by-word target translations as the source and the original source sentences as the target. We select $k$ instances from $D_{w2w}$, along with their corresponding original sentences from $D_T$. These selected sentence pairs will serve as ICL examples in the next step to translate sentences in $D_U$ from $L_t$ to $L_s$ using the LLM, thereby achieving more natural translations than word-by-word ones.

\subsubsection{$L_s$ to $L_t$ ICL Example Mining} \label{sec:unsuper_mining}

Following the back-translation step, we acquire a set of sentence pairs that both have natural language and that correct the word-by-word translation word order and grammar issues. Our next step involves using these pairs for ICL to translate sentences in the test set $D_T$. For that, we propose \textbf{TopK+BM25}, a method to select the optimal $k$ examples from these generated pairs. Specifically, for each test example in $D_T$, we identify the most relevant sentence pairs based on two criteria: (i) selecting pairs where the cosine similarity score (computed using $sim(.,.)$) between the source and target sentences exceeds a specified threshold $\tau$, and (ii) from these selected pairs, choosing the $k$ pairs that yield the highest BM25 ~\cite{10.1145/2682862.2682863} similarity score for each test example.

\iffalse
For that, we propose two methods to select $k$ examples from these generated pairs: 
\begin{enumerate}

    \item \textbf{Using Cosine Similarity.} The mined sentence pairs are sorted in descending order based on the cosine similarity score between each source and target in the collection, where the cosine similarity is calculated using the $sim(.,.)$ function. Subsequently, the top $k$ examples are selected to be the in-context examples used to translate the input sentences in $D_T$.

    \item \textbf{Using BM25~\cite{10.1145/2682862.2682863}.} To further refine our selection, we employ a more targeted approach in choosing in-context examples. Specifically, for each test example in the test dataset $D_T$, we identify the most relevant sentence pairs based on two criteria: (i) selecting pairs where the cosine similarity score between the source and target sentences exceeds a specified threshold $\tau$, and (ii) from these selected pairs, choosing the $k$ pairs that yield the highest BM25 similarity score for each test example.

\end{enumerate}
\fi

%These steps~\ref{sec:tran_step_1} and~\ref{sec:unsuper_mining} can be iterated multiple times to yield higher-quality $k$ in-context examples.

\section{Experimental Setup}

\paragraph{Models.} We perform our experiments using two different models, namely Bloom (7B)~\cite{workshop2023bloom} and Llama-3 (8B)~\cite{llama3modelcard}.%, which are decoder-only language models. %We selected 7-8B models due to computational resource constraints.
 We use their base model versions for all the experiments. For the similarity function $sim(x,y)$, we employ an XLM-R \cite{conneau-etal-2020-unsupervised} version of Sentence-BERT~\cite{reimers-gurevych-2019-sentence} to compute embeddings for input texts. We then use cosine similarity to measure the similarity score between embeddings of two sentences.

\paragraph{Datasets.} We conduct our evaluation on the FLORES-200 dataset~\cite{nllbteam2022language}, which covers 200 languages.\footnote{https://github.com/openlanguagedata/flores} Each language has a test set of 1,012  examples (we use it as our test set $D_T$) and a validation set of 997 examples (we use it as our unlabeled sentences $D_U$). We filter out these languages to end up with 64 languages (listed in Table~\ref{tab:covered_langs} in Appendix~\ref{app:data}) on which we evaluate our approach. For filtering, we keep languages for which we have access to a list of vocabularies ($V_s$ and $V_t$ from Section~\ref{sec:method}). From our 64 languages, we create 288 directions by forming different language pair combinations. Our combinations cover four main scenarios: (i) \textit{English-centric} experiments, with English as source or as target; (ii) \textit{language-family-centric} experiments (12 language families); (iii) \textit{script-centric} experiments (14 language scripts), and (iv) \textit{resource-level-centric} experiments (from very low-resource to high-resource languages).
 %For the lists of vocabularies per language, $V_s$ and $V_t$, we use the static word embedding FastText models developed by~\citet{grave-etal-2018-learning} for 157 languages. Each model for each language has a vocabulary of words sorted by frequency and covers up to 200,000 words.\footnote{In our work, we select only the top 10,000 most frequent words for each language.}
For vocabularies per language, $V_s$ and $V_t$, we utilize FastText 
embedding models~\cite{grave-etal-2018-learning}. These models contain up to 200,000 frequency-sorted words per language; we select the top 10,000 for each language.

\paragraph{Baselines and Comparisons.} We compare our method to a wide range of baselines involving ICL, as follows (comparisons to UMT SOTA in \ref{sec:analy}):

\begin{itemize}[noitemsep,topsep=0pt]
\item[\textbullet] \textbf{\textit{Zero}-Shot}: The LLM generates translations without any examples.
\item[\textbullet] \textbf{$k$-Shot (Regular ICL)}: The LLM uses $k$ random example translations from the validation set as prompts.
\item[\textbullet] \textbf{$k$-Shot (Regular BM25 ICL)}: The LLM uses $k$ example translations from the validation set as prompts. These examples are selected specifically for each input using BM25.
\item[\textbullet] \textbf{Unsupervised Word-by-Word (UW2W)}: Translates individual words using the word translation model from Section~\ref{sec:w2w}.
\item[\textbullet] \textbf{Unsupervised $k$-Shot (Random)}: Uses $k$ randomly sampled mined sentence pairs (Section~\ref{sec:unsuper_mining}) as prompts.
\item[\textbullet] \textbf{Unsupervised $k$-Shot (TopK)}: Employs $k$ mined sentence pairs with the highest cosine similarity scores (Section~\ref{sec:unsuper_mining}) as prompts.
\item[\textbullet] \textbf{Unsupervised $k$-Shot ICL (TopK+BM25)}: Selects the top $k$ mined sentence pairs sorted by cosine similarity and BM25 (Section~\ref{sec:unsuper_mining}) as prompts.
\end{itemize}

For simplicity and reproducibility of the random selection, we select the first $k$ examples and consider them as randomly sampled.

\paragraph{Evaluation Metrics.} We use two metrics that reflect performance across various levels, following~\citet{nllbteam2022language}: (i) spBLEU \cite{goyal-etal-2022-flores}, a variant of BLEU \cite{papineni-etal-2002-bleu}, and (ii) chrF++ \cite{popovic-2017-chrf}. spBLEU accounts for the morphological richness of languages by tokenizing sentences into subwords before computing the BLEU score, while chrF++ evaluates character n-gram matches and has been shown to correlate with human annotations, particularly for low-resource languages~\cite{popovic-2017-chrf}.

\paragraph{Implementation Details.} We use vLLM \cite{10.1145/3600006.3613165} for LLMs inference in our experiments. For the word translation phase, we set $k_{wp} = 10$ as the number of $k_{wp}$ word pairs to mine for ICL word translation; and for the iteration from source to target language, we generate ten word translations for each source word. For the sentence translation phase, we set $k = 8$ as the number of $k$ in-context sentence pair examples.\footnote{Based on the experiments conducted with various values of $k$ (presented in Figure \ref{fig:spbleu_shots_full} in \ref{sec:app_anal}), $k=8$ was the optimal.} Concerning the threshold $\tau$, we set it as $0.90$.\footnote{Experiments were conducted with $\tau$ values ranging from $0.0$ to $0.90$ (presented in Table \ref{tab:tau-mt-scores} in Appendix \ref{sec:app_anal}). Values between $0.8$ and $0.9$ yielded optimal spBLEU scores.} If this criterion is not met, we select the top $20$ pairs and apply BM25 selection from them. Our experiments are conducted on a server with two A100 (40GB) GPUs. %We set $4096$ tokens as the maximum sequence length for the LLM.
To make our inference more efficient, we used prefix caching as most of the prompt instructions are redundant.

%For the footnote
%Additionally, using larger values of $k$ may result in prompts that exceed the LLM context length.
\section{Results and Analysis}
\subsection{Results}

Table~\ref{tab:main_results} shows a subset of our Llama-3 experimental results (full results in Table~\ref{tab:full_results}, Appendix \ref{sec:app_results}). It focuses on language pairs with English as the source or target, covering four resource levels: high, medium, low, and very low. Three non-English languages per level are randomly selected. Our unsupervised TopK+BM25 method achieves an average spBLEU of 55.76 on this subset, outperforming the k-shot regular ICL (55.07 spBLEU) and on par with the k-shot regular BM25 ICL (56.93 spBLEU), without relying on human-annotated data.

Additionally, the results indicate that our \textit{UW2W} translation method ($40.82$ spBLEU) outperforms \textit{zero}-shot MT ($37.79$ spBLEU). Furthermore, our \textit{TopK+BM25} selection algorithm further enhances the results of these \textit{UW2W} translations compared to the \textit{Random ICL} and \textit{TopK ICL}. For instance, in the English to Italian (\textit{eng\_Latn → ita\_Latn}) translation, our \textit{TopK+BM25} achieves a $68.34$ spBLEU. This approach outperforms the UW2W method, \textit{Random ICL}, \textit{TopK ICL}, and the $k$-shot regular BM25 method, which achieve $54.77$, $66.91$, $66.57$, and $64.70$ spBLEU scores, respectively.

The full results for Llama-3 in Table \ref{tab:full_results} in Appendix~\ref{sec:app_results} show that our approach achieves an average spBLEU of $41.75$ and a chrF++ of $28.30$. In comparison, the regular BM25-Based $k$-shot method achieves $42.19$ spBLEU and $28.89$ chrF++. For Bloom, the results are provided in Table \ref{tab:full_results_bloom} in Appendix~\ref{sec:app_results}. Consistent with the trend observed in Llama-3, Bloom achieves an average spBLEU score of $25.34$ and a chrF++ score of $15.29$. These results surpass those of the regular $k$-shot, which achieves an average spBLEU of $23.17$ and a chrF++ of $14.95$.

%Also, we provide a comparison of our UMT method to existing UMT methods such as the SMT+UMT method from \citet{artetxe-etal-2019-effective} and \citet{lample-etal-2018-phrase}. Results presented in Table \ref{tab:wmt_compare} in Appendix \ref{sec:app_anal} show that LLMs acquire strong \textit{Zero}-shot multilingual capabilities that can outperform existing UMT methods, while our TopK+BM25 method boosts these capabilities further.

%When applying a similarity score to select the most confident in-context examples, our unsupervised ICL score is boosted to $39.94$ spBLEU. Moreover, when we use BM25 similarity to obtain input-specific examples from these most confident in-context examples, the performance further improves to $40.78$ spBLEU. Our approach has achieved similar performance for the other language resource groups. Additionally, the results show that the performance of the 1-shot setting is better than the 8-shot setting in the supervised ICL for the majority of cases. We attribute this to the fact that LLMs perform well when the downstream task involves the English language, as evidenced by the consistently high spBLEU scores when English is the target language.

%\begin{tabularx}{\textwidth}{l l c r r r | *{4}{>{\centering\arraybackslash}X}}

% \begin{tabularx}{\textwidth}{l l c *{7}{>{\centering\arraybackslash}X}}
%\begin{tabularx}{\textwidth}{l l c r r r | *{4}{r}}

\begin{table*}[t]
\centering
\tiny
\begin{tabularx}{\textwidth}{l l c | *{3}{>{\centering\arraybackslash}X} | *{4}{>{\centering\arraybackslash}X}}
\toprule
\multirow{2}{*}{\textbf{Resource Level}} & \multirow{2}{*}{\textbf{Language}} & \multirow{2}{*}{\textbf{Direction}} & \multicolumn{3}{c|}{\textbf{Baselines}} & \multicolumn{4}{c}{\textbf{Ours (Unsupervised)}} \\
& & & \textbf{\textit{Zero}-Shot} & \textbf{\textit{k}-Shot} & \textbf{BM25} & \textbf{UW2W} & \textbf{Random} & \textbf{TopK} & \textbf{TopK+BM25} \\
\midrule
\multirow{6}{*}{\textbf{High}} 
& \multirow{2}{*}{\textbf{ita\_Latn}} & $\rightarrow$ & 56.64 & 64.02 & 64.70 & 54.77 & 66.91 & 66.57 & \textbf{68.34} \\
& & $\leftarrow$ & 55.48 & 71.55 & 73.10 & 49.79 & 73.63 & 70.51 & \textbf{73.91} \\
\addlinespace[1mm]
& \multirow{2}{*}{\textbf{rus\_Cyrl}} & $\rightarrow$ & 41.95 & 56.81 & 60.76 & 45.98 & 58.54 & 59.64 & \textbf{63.47} \\
& & $\leftarrow$ & 55.15 & 68.93 & 71.53 & 34.09 & 72.09 & 71.81 & \textbf{74.38} \\
\addlinespace[1mm]
& \multirow{2}{*}{\textbf{kor\_Hang}} & $\rightarrow$ & 22.32 & 39.77 & 40.47 & 28.94 & 37.68 & 40.94 & \textbf{41.23} \\
& & $\leftarrow$ & 37.08 & 60.61 & 63.73 & 49.51 & 61.94 & 41.19 & \textbf{67.74} \\
\midrule
\multirow{6}{*}{\textbf{Medium}} 
& \multirow{2}{*}{\textbf{arb\_Arab}} & $\rightarrow$ & 13.28 & 33.11 & 38.40 & \textbf{48.23} & 32.01 & 30.67 & 36.85 \\
& & $\leftarrow$ & 43.19 & 66.92 &71.84 & 48.23 & 66.34 & 72.53 & \textbf{74.35} \\
\addlinespace[1mm]
& \multirow{2}{*}{\textbf{nld\_Latn}} & $\rightarrow$ & 55.75 & 65.10 & 63.08 & 57.46 & 66.98 & 67.30 & \textbf{67.86} \\
& & $\leftarrow$ & 63.24 & 70.75 & 69.83 & 56.95 & 71.94 & \textbf{73.91} & 73.36 \\
\addlinespace[1mm]
& \multirow{2}{*}{\textbf{cat\_Latn}} & $\rightarrow$ & 60.83 & 69.05 & 70.84 & 57.92 & 69.90 & 69.75 & \textbf{72.95} \\
& & $\leftarrow$ & 53.65 & 76.82 & 78.96 & 55.52 & 77.62 & 77.40 & \textbf{80.28} \\
\midrule
\multirow{6}{*}{\textbf{Low}} 
& \multirow{2}{*}{\textbf{zsm\_Latn}} & $\rightarrow$ & 53.66 & 68.85 & 70.88 & 57.41 & 69.82 & 70.55 & \textbf{71.20} \\
& & $\leftarrow$ & 54.11 & 73.96 & 75.83 & 57.25 & 70.17 & 71.11 & \textbf{77.07} \\
\addlinespace[1mm]
& \multirow{2}{*}{\textbf{amh\_Ethi}} & $\rightarrow$ & 4.72 & 14.48 & \textbf{17.62} & 7.23 & 12.43 & 4.92 & 5.50 \\
& & $\leftarrow$ & 18.94 & 34.52 & \textbf{42.68} & 20.99 & 33.13 & 12.32 & 21.48 \\
\addlinespace[1mm]
& \multirow{2}{*}{\textbf{som\_Latn}} & $\rightarrow$ & 9.02 & 26.31 & \textbf{32.10} & 29.64 & 22.35 & 23.03 & 23.02 \\
& & $\leftarrow$ & 22.30 & 35.44 & \textbf{40.37} & 33.25 & 31.07 & 26.12 & 34.23 \\
\midrule
\multirow{6}{*}{\textbf{Very Low}} 
& \multirow{2}{*}{\textbf{bel\_Cyrl}} & $\rightarrow$ & 17.19 & 36.40 & 37.76 & 29.40 & 43.50 & 42.17 & \textbf{45.39} \\
& & $\leftarrow$ & 46.66 & 60.52 & 62.47 & 38.16 & \textbf{65.18} & 57.05 & 65.08 \\
\addlinespace[1mm]
& \multirow{2}{*}{\textbf{asm\_Beng}} & $\rightarrow$ & 8.16 & 34.79 & \textbf{37.29} & 5.24 & 28.88 & 5.42 & 5.54 \\
& & $\leftarrow$ & 26.14 & 54.26 & \textbf{59.31} & 6.22 & 45.63 & 32.28 & 50.92 \\
\addlinespace[1mm]
& \multirow{2}{*}{\textbf{oci\_Latn}} & $\rightarrow$ & 26.96 & 58.37 & \textbf{62.22} & 53.25 & 59.61 & 61.83 & 61.91 \\
& & $\leftarrow$ & 60.48 & 80.42 & \textbf{82.58} & 54.38 & 81.43 & 63.83 & 82.13 \\
\midrule
\textbf{Average} & & & 37.79 & 55.07 & \textbf{56.93} & 40.82 & 54.95 & 50.54 & 55.76 \\
\bottomrule
\end{tabularx}
\caption{spBLEU scores for a subset of language pairs involving English as either the source (→) or target (←) language, using the Flores-200 dataset across various resource levels. The results compare our UMT approaches to Zero-Shot and regular $k$-shot baselines using Llama-3. The highest scores for each language pair are highlighted in bold. The "Average" row displays the mean score for all included pairs. (Full results in Table~\ref{tab:full_results} in Appendix~\ref{sec:app_results}).}
\label{tab:main_results}
\end{table*}

\subsection{Analysis} \label{sec:analy}
\paragraph{How does our TopK+BM25 compare to existing UMT approaches?} Using the WMT-14 and WMT-16 benchmarks covering English-French and English-German, respectively, we directly compare our TopK+BM25 (with LLaMA-3) to state-of-the-art UMT methods—including Statistical Machine Translation (SMT), Neural Machine Translation (NMT), and hybrid SMT+NMT methods—as well as strong UMT baselines built on top of XLM (see Table \ref{tab:wmt_compare}).  On average, our method achieves a BLEU score of $40.13$, outperforming all the previous UMT baselines including the XLM-based ones by an average of $7$ BLEU score points. The best baseline is XLM with Online Self-Training followed by self-correction \cite{lu-zhang-2024-improving}, which achieves an average BLEU score of $33.68$. Specifically, the same trend is observed for all pairs that we evaluate. This shows the strength of our method compared to leading UMT models. 

%For English→French translation, our method achieves a BLEU score of $36.62$, outperforming the best SMT+NMT baseline's $33.60$ and zero-shot LLaMA-3's $17.93$. Similarly, for French→English, it attains $37.13$, surpassing the best baseline UMT method ($33.20$) and zero-shot performance ($23.13$). The same trend of results is achieved for the English-German pair from WMT-16. It is worth noting that all the other UMT methods involve training, while ours rely only on ICL. %These results demonstrate the effectiveness of our TopK+BM25 approach for UMT, indicating that leveraging LLMs with our unsupervised techniques significantly improves translation quality.

% Include baselines from https://aclanthology.org/2024.lrec-main.783.pdf

\begin{table}[htbp]
\centering
\resizebox{0.48\textwidth}{!}{%
\begin{tabular}{l l| c c| c c| c}
\toprule
\multirow{2}{*}{\textbf{Type}} & \multirow{2}{*}{\textbf{Method}} & \multicolumn{2}{c|}{\textbf{WMT-14}} & \multicolumn{2}{c|}{\textbf{WMT-16}} & \multirow{2}{*}{\textbf{Avg}} \\
\cline{3-6}
& & \multicolumn{1}{c}{\textbf{En$\rightarrow$Fr}} & \multicolumn{1}{c|}{\textbf{Fr$\rightarrow$En}} & \multicolumn{1}{c}{\textbf{En$\rightarrow$De}} & \multicolumn{1}{c|}{\textbf{De$\rightarrow$En}} & \multicolumn{1}{c}{} \\
\midrule
\multirow{1}{*}{NMT} & \citet{lample-etal-2018-phrase} & 25.10 & 24.20 & 17.20 & 21.00 & 21.88 \\
\midrule
\multirow{2}{*}{SMT} & \citet{lample-etal-2018-phrase} & 28.10 & 27.20 & 17.90 & 22.90 & 24.03 \\
& \citet{artetxe-etal-2019-effective} & 27.80 & 27.90 & 19.40 & 24.80 & 24.98 \\
\midrule
\multirow{2}{*}{SMT+NMT} & \citet{lample-etal-2018-phrase} & 27.60 & 27.70 & 20.20 & 25.20 & 25.18 \\
& \citet{artetxe-etal-2019-effective} & 33.60 & 33.20 & 26.40 & 33.80 & 31.75 \\
\midrule
\multirow{2}{*}{XLM} & \citet{he-etal-2022-bridging} & 36.40 & 34.30 & 28.30 & 34.50 & 33.38 \\
& \citet{lu-zhang-2024-improving} & 36.90 & 34.70 & 28.30 & 34.80 & 33.68 \\
\midrule
Ours & TopK+BM25 & {\bfseries42.47} & {\bfseries39.78} & {\bfseries35.04} & {\bfseries43.26} & {\bfseries40.13} \\
\bottomrule
\end{tabular}
}
\caption{Comparison of our UMT method (using Beam Search Decoding) against existing UMT methods relying on NMT and SMT. The baseline results presented in the table are BLEU scores reported in their paper. Our BLEU scores were computed using the \texttt{mteval-v13a.pl} script.}
\label{tab:wmt_compare}
\end{table}

\paragraph{How does our TopK+BM25 compare to out-of-domain (OOD) $k$-shot?} To simulate a more realistic use case of our method. We evaluate the performance achieved using our TopK+BM25 method when applied on unlabeled in-domain texts and compare it to regular $k$-shot when the shots are coming from out-of-domain (OOD) pairs. The parallel texts used in this experiment come from the following datasets: Flores-200, OPUS-100 \cite{zhang-etal-2020-improving}, WMT-16, and WebCrawl African \cite{vegi-etal-2022-webcrawl}.
Table \ref{tab:odd-exps} presents the results of this experiment. Results show that our method consistently outperforms the baseline shots in the majority of performed experiments. For example, when the test data is from Flores Dataset and OOD is from WebCrawl African, the performance when choosing in-domain shots for Yoruba to English is 39.29 spBLEU, while OOD shots result in 34.90, widely underperforming our unsupervised method which achieves 40.54 spBLEU. The same scenario applies to the opposite direction, English to Yoruba, where the in-domain shots achieve 26.23 spBLEU, outperforming our unsupervised method which achieves 22.63 spBLEU. However, our method still largely outperforms the OOD shots method, which achieves 14.10 spBLEU.

\begin{table}[htbp]
\centering
\resizebox{0.48\textwidth}{!}{%
\begin{tabular}{l|l|l|lll}
\hline
\textbf{Pair}                  & \textbf{Test Data}               & \textbf{OOD Data}                          & \textbf{ID shots} & \textbf{OOD shots} & \textbf{Ours}  \\ \hline
\textbf{asm → eng} & \multirow{9}{*}{Flores} & \multirow{9}{*}{OPUS}             & \textbf{54.26}    & 52.82     & 50.92 \\
\textbf{bel → eng} &                         &                                   & 60.52    & 63.57     & \textbf{65.08} \\
\textbf{bul → eng} &                         &                                   & 73.23    & 71.29     & \textbf{77.52} \\
\textbf{ben → eng} &                         &                                   & \textbf{62.15}    & 61.13     & 57.13 \\
\textbf{cat → eng} &                         &                                   & 76.82    & 75.99     & \textbf{80.28} \\
\textbf{deu → eng} &                         &                                   & 77.23    & 73.12     & \textbf{79.14} \\
\textbf{ell → eng} &                         &                                   & 72.13    & 68.59     & \textbf{73.86} \\
\textbf{eng → gle} &                         &                                   & 38.00    & 30.21     & \textbf{38.90} \\ \hline
\textbf{yor → eng} & \multirow{2}{*}{Flores} & \multirow{2}{*}{WebCrawl African} & 39.29    & 34.90     & \textbf{40.54} \\
\textbf{eng → yor} &                         &                                   & \textbf{26.23}    & 14.10     & 22.63 \\ \hline
\textbf{eng → deu} & Flores                  & WMT-16                            & 64.99    & 60.34     & \textbf{67.82} \\ \hline
\textbf{eng → deu} & WMT-16                  & Flores                            & 64.37    & 62.31     & \textbf{68.03} \\ \hline
\end{tabular}}
\caption{spBLEU scores of our UMT method compared to in-domain (ID) and out-of-domain (OOD) regular $k$-shot.}
\label{tab:odd-exps}
\end{table}

\begin{figure}[] 
    \centering
    \includegraphics[width=0.5\textwidth, height=0.25\textheight]{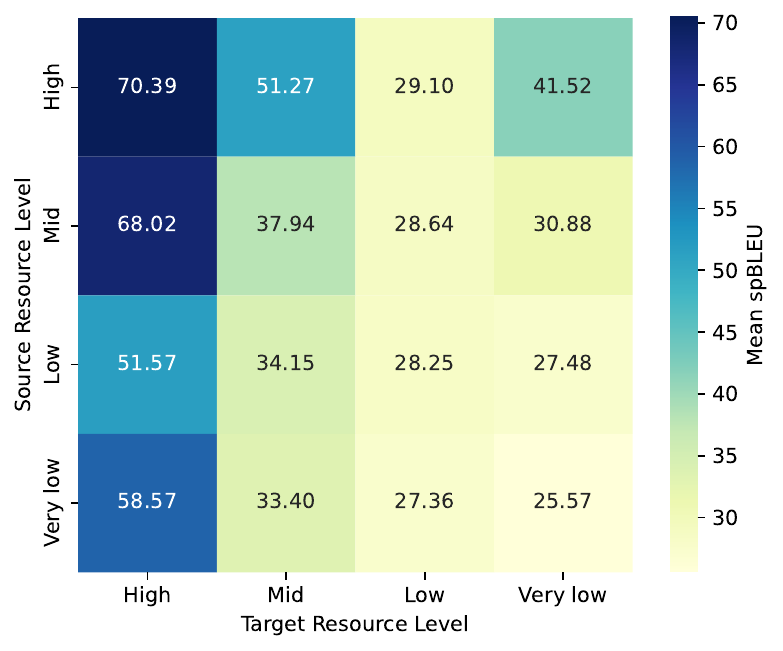} 
    \caption{Mean spBLEU TopK+BM25 scores using Llama-3, showing MT performance across resource levels. (Bloom results in Appendix \ref{sec:app_anal}, Figure \ref{fig:spbleu_res_lev_agg_bloom}).} 
    \label{fig:spbleu_res_lev_agg}
\end{figure}

\paragraph{Does the language resource level affect the UMT performance?} Figure \ref{fig:spbleu_res_lev_agg} shows the mean spBLEU scores from the TopK+BM25 experiments by language resource level. The results demonstrate that the performance of our unsupervised approach is influenced by both the source and target language resource levels, with generally higher scores when the \textit{target} language has high resource levels. Notably, the average spBLEU is highest when both languages are high-resource, reaching a score of $70.39$. Conversely, performance decreases as the resource level of the target language diminishes. Unsurprisingly, the lowest scores are observed when both source and target are very low-resource ($25.57$ spBLEU). This trend highlights the impact of resource level on MT performance.%, emphasizing the advantage of high resources.% languages in achieving better translation results.

\paragraph{Does the language family and script affect ICL performance in UMT?} Tables \ref{tab:lng_fam_eff} and \ref{tab:lng_scr_eff} present the average spBLEU scores for each language family and script when used as the target language in our approach (TopK+BM25), respectively. Results from the language family experiments indicate high variance in performance among different language families. For instance, the performance is optimal for Romance and Germanic languages. These groups comprise most of the Indo-European languages, which are generally well-resourced. Meanwhile, performance declines when the target languages belong to the Nilotic+Other AC or Indo-Aryan language families. These families primarily include African and South Asian languages, many of which are considered low to very-low resourced. Similarly, performance per script results also show high variance, where languages written in Latin script have a better average spBLEU and under-represented scripts such as Ge'ez and Bengali have lower scores.

%Table \ref{tab:lang_family_results} presents the average aggregated results of experiments per language family. The aggregation was performed by computing the mean of spBLEU scores for each language family when represented as source or target language. 

\begin{table}[!htp]\centering
\tiny
\resizebox{0.48\textwidth}{!}{%
\begin{tabular}{cccc}\toprule
\textbf{Afro-Asiatic} &\textbf{Austronesian} &\textbf{Balto-Slavic} \\
25.90 &49.50 &50.30 \\\toprule
\textbf{Bantu} &\textbf{Germanic} &\textbf{Indo-Aryan} \\
24.60 &63.00 &23.10 \\\toprule
\textbf{Nilotic+Other AC} &\textbf{Romance} &\textbf{Turkic} \\
17.30 &60.60 &31.90 \\\toprule
\textbf{Uralic} & & \\
45.60 & & \\
\bottomrule
\end{tabular}}
\caption{Mean spBLEU TopK+BM25 scores using Llama-3, showing MT performance across language families (as targets).}\label{tab:lng_fam_eff}

\end{table}

\begin{table}[!htp]\centering
\tiny
\resizebox{0.48\textwidth}{!}{%
\begin{tabular}{ccccc}\toprule
\textbf{Arabic} &\textbf{Armenian} &\textbf{Bengali} &\textbf{Cyrillic} \\
24.00 &30.60 &21.00 &34.90 \\\toprule
\textbf{Devanagari} &\textbf{Georgian} &\textbf{Ge'ez} &\textbf{Greek} \\
27.70 &23.60 &5.30 &33.20 \\\toprule
\textbf{Han, Hiragana, Katakana} &\textbf{Hangul} &\textbf{Hebrew} &\textbf{Latin} \\
37.20 &24.90 &51.10 &51.20 \\\toprule
\textbf{Perso-Arabic} &\textbf{Thai} & & \\
23.80 &43.20 & & \\
\bottomrule
\end{tabular}}
\caption{Mean spBLEU TopK+BM25 scores using Llama-3, showing MT performance across language scripts (as targets).}\label{tab:lng_scr_eff}
\end{table}

%\mam stopped here

%\paragraph{\textcolor{red}{How does our method compares to existing UMT approaches?}}

\paragraph{Which linguistic features impact UMT performance?} We analyze the influence of several linguistic features on our UMT performance. These features are categorized as follows:

\noindent \textbf{LLM-Dependent:} \textbf{(i) UW2W spBLEU:} The performance of the UW2W translation, reflecting the importance of the word-level translation phase in our method. \textbf{(ii) Subword Overlap:} After tokenization, the rate of subwords shared between the source and target languages' parallel data. \textbf{ (iii)Target Language Compression Ratio (TLCR):} Reflects the compression ratio of the target language and addresses the over-segmentation issue in morphologically rich languages.

\noindent \textbf{LLM-Independent:}
\textbf{(i) Word Overlap:} The rate of words shared between the source and target languages' parallel data. \textbf{(ii) Linguistic Distances:} Four distances from \citet{lin-etal-2019-choosing}, queried from the URIEL Typological Database \cite{littell-etal-2017-uriel}: \textbf{\textit{Geographic Distance:}} The orthodromic distance between languages on Earth's surface, divided by the antipodal distance, \textbf{\textit{Genetic Distance:}} The genealogical distance between languages, \textbf{\textit{Syntactic Distance:}} The cosine distance between feature vectors derived from the syntactic structures of the languages, and \textbf{\textit{Phonological Distance:}} The cosine distance between phonological feature vectors of the languages.

Table \ref{tab:linguistic_features} presents Pearson correlation coefficients between these features and UMT scores using the TopK+BM25 approach. Results are grouped by resource level (RL): high and medium (HM), and low and very low (LVL). Since both Bloom and LLaMa-3 models show similar trends, we focus on LLaMa-3. UW2W spBLEU exhibits the strongest positive correlation with UMT performance (0.71 for HM, 0.78 for LVL), underscoring the importance of the initial word-level translation phase. Subword overlap shows a moderate positive correlation (0.13 for HM, 0.39 for LVL), indicating that languages sharing more subwords achieve better UMT performance. Word overlap also demonstrates a positive correlation (0.37 for HM, 0.57 for LVL), suggesting that shared vocabulary improves UMT tasks. The target language compression ratio (TL~CR) has a negative correlation (-0.37 for HM, -0.37 for LVL), implying better UMT performance for languages with less complex morphology or better tokenizer vocabulary coverage.

Among linguistic distances, genetic distance shows the strongest correlation, especially for low-resource languages (-0.42 for LVL), while its impact is less pronounced for high and medium-resource languages (-0.21 for HM). Syntactic distance has a moderate negative correlation for high-resource languages (-0.36 for HM) but is not significant for low-resource languages (-0.13 for LVL). Other distances—geographic, and phonological—exhibit weak or insignificant correlations with UMT.

\begin{table}[htbp]
\centering
\small
\begin{adjustbox}{max width=0.48\textwidth}
\begin{tabular}{@{}llccc@{}}
\toprule
& \textbf{Feature} & \textbf{RL} & \textbf{Bloom} & \textbf{LLaMa-3} \\
\midrule
\multirow{6}{*}{\rotatebox[origin=c]{90}{\textbf{LLM-Dependent}}} 
 & \multirow{2}{*}{\textbf{UW2W spBLEU}} & HM & 0.85 (0.0) & 0.71 (0.0) \\
 & & LVL & 0.89 (0.0) & 0.78 (0.0) \\ \cmidrule(l){2-5}
 & \multirow{2}{*}{\textbf{Subword Overlap}} & HM & 0.18 (0.03) & 0.13 (0.09) \\
 & & LVL & 0.54 (0.0) & 0.39 (0.0) \\ \cmidrule(l){2-5}
 & \multirow{2}{*}{\textbf{TL CR}} & HM & -0.31 (0.0) & -0.37 (0.0) \\
 & & LVL & -0.62 (0.0) & -0.37 (0.0) \\ 
\midrule
\multirow{12}{*}{\rotatebox[origin=c]{90}{\textbf{LLM-Independent}}} 
 & \multirow{2}{*}{\textbf{Word Overlap}} & HM & 0.6 (0.0) & 0.37 (0.0) \\
 & & LVL & 0.81 (0.0) & 0.57 (0.0) \\ \cmidrule(l){2-5}
 & \multirow{2}{*}{\textbf{Geographic}} & HM & -0.19 (0.02) & -0.17 (0.02) \\
 & & LVL & -0.23 (0.11) & -0.2 (0.12) \\ \cmidrule(l){2-5}
 & \multirow{2}{*}{\textbf{Genetic}} & HM & -0.34 (0.0) & -0.21 (0.0) \\
 & & LVL & -0.4 (0.0) & -0.42 (0.0) \\ \cmidrule(l){2-5}

 & \multirow{2}{*}{\textbf{Syntactic}} & HM & -0.18 (0.04) & -0.36 (0.0) \\
 & & LVL & -0.18 (0.21) & -0.13 (0.31) \\ \cmidrule(l){2-5}
 & \multirow{2}{*}{\textbf{Phonological}} & HM & 0.02 (0.77) & -0.16 (0.03) \\
 & & LVL & 0.0 (0.99) & -0.08 (0.51) \\
\bottomrule
\end{tabular}
\end{adjustbox}
\caption{Pearson correlation (p-value) values of our UMT model spBLEU score using Bloom and LLaMa-3 and different linguistic features.}
\label{tab:linguistic_features}
\end{table}

\section{Conclusion}

In this work, we proposed an unsupervised approach to mine in-context examples for machine translation, enabling effective translation across a wide range of languages, including different language families, scripts, and resource-level settings. Our method combines word-level and sentence-level translation, along with a filtering criterion to select optimal examples using multilingual sentence embedding similarity and BM25. Evaluations using Llama-3 and Bloom on 288 directions from Flores-200 demonstrated that our approach achieved better or comparable performance to regular MT ICL, highlighting its potential to overcome the limitations of access to parallel examples in various language pairs. Our analysis shows the impact of several factors such as resource level, script, language family, and different linguistic distances on the performance of our approach. This work showcases the effectiveness of mining in-context examples for improved translation performance especially for under-represented languages, opening up new possibilities for UMT.

\section*{Limitations}
While our unsupervised approach demonstrates competitive performance compared to the regular baselines, it is important to highlight the following limitations:
\begin{itemize}
    \item \textbf{Dependency on LLM Language Coverage:} The results presented in this paper indicate that the performance of both regular and unsupervised $k$-shot ICL for MT is predominantly correlated with the resource level of the target language. This correlation poses significant challenges in developing reliable MT systems for low-resource languages. Primarily, the lack of extensive pre-training corpora for these languages results in poorer generation performance compared to that for high-resource languages. Our main focus in this work was to achieve MT performance comparable to the $k$-shot performance without the need for human-annotated shots.

    \item \textbf{Dependency on Multilingual Embedding Representation for Filtering.} Although our TopK+BM25 algorithm demonstrates effective performance in selecting semantically similar pairs using the similarity function $sim(x,y)$, it depends on XLM-R embeddings. XLM-R does not support all languages, resulting in noisy and inaccurate representations for several languages, particularly those that are low-resource.
 
    \item \textbf{Time Complexity for Unsupervised Example Mining:} Our unsupervised method for example mining employs a two-step process—first identifying in-context word pairs and then finding in-context sentence pairs—which requires more computational effort than the regular method. However, our method is computationally efficient compared to other UMT methods~\cite{artetxe-etal-2019-effective,lu-zhang-2024-improving}. We also use efficient techniques such as prefix-caching and random sampling (instead of beam search) on two A100 (40GB) GPUs to optimize performance and minimize processing time.   %This long duration makes it hard to adjust the hyperparameters like how many words to translate in the word-level translation phase, how many translations to create for each word, and how to choose the best pairs of sentences to filter. Also, it's difficult to use bigger models with 13B or 70B parameters because they need more resources.

    \item \textbf{Challenges in Optimizing Prompt Templates and In-Context Example Ordering:} The effectiveness of ICL is influenced by the design of prompt templates and the structure and ordering of in-context examples. Although we follow the best practices established in previous work~\cite{10.5555/3618408.3620130}, identifying the optimal combination for each language pair and translation direction remains a significant challenge.

\end{itemize}

\section*{Ethics Statement}

This research aims to enhance language technology by addressing lexical disparities among languages, groups, and cultures. We focus on utilizing LLMs through the unsupervised mining of in-context sentence pairs, which facilitates sentence-level translation across languages.

Our study includes 64 languages, representing 288 language pairs from 12 language families and 14 scripts, across various resource levels. The goal is to extend unsupervised machine translation to these languages using zero-shot techniques. Ultimately, this work seeks to increase access to technology for diverse populations.

The dataset employed in our research, Flores-200, is publicly available and, in our assessment, poses no risks. However, for any real-world application, we recommend conducting thorough evaluations and analyses before deployment.

\section*{Acknowledgments}\label{sec:acknow}
We acknowledge support from Canada Research Chairs (CRC), the Natural Sciences and Engineering Research Council of Canada (NSERC; RGPIN-2018-04267, RGPIN-2020-05408), the Social Sciences and Humanities Research Council of Canada (SSHRC; 895-2020-1004; 895-2021-1008), Canadian Foundation for Innovation (CFI; 37771), Digital Research Alliance of Canada,\footnote{\href{https://alliancecan.ca}{https://alliancecan.ca}} and UBC Advanced Research Computing-Sockeye.\footnote{\href{https://arc.ubc.ca/ubc-arc-sockeye}{https://arc.ubc.ca/ubc-arc-sockeye}}

\bibliography{anthology,custom}
%\bibliography{custom}

\appendix

\section{Experimental Setup}
\subsection{Datasets} \label{app:data}

\begin{table*}[!htp]\centering
\scriptsize
\begin{tabular}{lrrrrrrr}\toprule
\textbf{Lang ID} &\textbf{ISO 639-3} &\textbf{Language} &\textbf{Family} &\textbf{Subgrouping} &\textbf{Script} &\textbf{Resource Level} \\\midrule
\textbf{fra\_Latn} &fra &French &Indo-European &Romance &Latin &High \\
\textbf{deu\_Latn} &deu &German &Indo-European &Germanic &Latin &High \\
\textbf{ita\_Latn} &ita &Italian &Indo-European &Romance &Latin &High \\
\textbf{por\_Latn} &por &Portuguese (Brazil) &Indo-European &Romance &Latin &High \\
\textbf{rus\_Cyrl} &rus &Russian &Indo-European &Balto-Slavic &Cyrillic &High \\
\textbf{spa\_Latn} &spa &Spanish (Latin America) &Indo-European &Romance &Latin &High \\
\textbf{eng\_Latn} &eng &English (Latin America) &Indo-European &Germanic &Latin &High \\
\bottomrule
\end{tabular}
\caption{The list of languages included in our experiments features several attributes for each language: its corresponding Lang ID from Flores-200, ISO 639-3 code, language name, language family, subgrouping, script, and resource level. This metadata was extracted from \citet{goyal-etal-2022-flores}.}\label{tab:covered_langs}

\end{table*}

\section{Results and Analysis}
\subsection{Results} \label{sec:app_results}

%\begin{table*}
%\scriptsize
%%

\end{xtabular}

\vspace{500pt}
\tablecaption{Detailed evaluation of translation scores for 222 language directions from the Flores-200 dataset using the Bloom model. Results include Zero-Shot and regular $k$-shot baselines, along with our unsupervised method employing various selection approaches. Scores for each language pair are presented as chrF++/spBLEU, with the highest scores highlighted in bold. The 'average' row indicates the mean score across all language pairs.}\label{tab:full_results_bloom}
\tablefirsthead{
\toprule
&\multicolumn{2}{c}{\parbox{4cm}{\centering \textbf{Baselines} \\ \textbf{(Zero-shot + Regular ICL)}}} &\multicolumn{4}{c}{\parbox{4cm}{\centering \textbf{Ours} \\ \textbf{(Unsupervised ICL)}}} \\\cmidrule{2-7}
\textbf{Pair} &\textbf{\textit{Zero}-Shot} &\textbf{$k$-Shot} &\textbf{UW2W} 
&\textbf{Random} &\textbf{TopK} &\textbf{TopK+BM25} \\\midrule
}

\tablehead{%
\toprule
&\multicolumn{2}{c}{\parbox{4cm}{\centering \textbf{Baselines} \\ \textbf{(Zero-shot + Regular ICL)}}} &\multicolumn{4}{c}{\parbox{4cm}{\centering \textbf{Ours} \\ \textbf{(Unsupervised ICL)}}} \\\cmidrule{2-7}
\textbf{Pair} &\textbf{\textit{Zero}-Shot} &\textbf{$k$-Shot} &\textbf{UW2W} 
&\textbf{Random} &\textbf{TopK} &\textbf{TopK+BM25} \\\midrule
}

\begin{xtabular}{lrr|rrrr}
\textbf{afr\_Latn → pol\_Latn} &10.25 / 12.93 &11.42 / 15.29 &\textbf{17.79} / \textbf{29.78} &11.54 / 18.14 &14.71 / 24.81 &14.17 / 23.61 \\
\textbf{pol\_Latn → afr\_Latn} &11.67 / 16.45 &12.49 / 18.24 &\textbf{16.67} / \textbf{29.78} &13.87 / 23.35 &15.02 / 25.52 &14.70 / 25.46 \\
\textbf{eng\_Latn → amh\_Ethi} &0.21 / 1.06 &2.83 / 7.35 &0.87 / 4.43 &2.11 / 5.34 &1.03 / 4.86 &1.06 / 4.89 \\
\textbf{amh\_Ethi → eng\_Latn} &0.77 / 0.00 &12.14 / 18.55 &10.55 / 17.84 &11.40 / 17.58 &11.72 / 19.91 &12.61 / 22.11 \\
\textbf{mlt\_Latn → amh\_Ethi} &0.38 / 2.55 &3.43 / 9.38 &0.79 / 3.80 &1.16 / 4.53 &0.85 / 3.86 &0.85 / 3.85 \\
\textbf{amh\_Ethi → mlt\_Latn} &4.17 / 2.84 &8.88 / 12.18 &7.67 / 11.78 &\textbf{10.06} / 15.76 &1.54 / 4.70 &4.80 / 9.27 \\
\textbf{tha\_Thai → amh\_Ethi} &0.19 / 1.55 &0.28 / 1.19 &\textbf{0.83} / \textbf{2.76} &0.18 / 0.46 &0.17 / 0.83 &0.26 / 1.46 \\
\textbf{amh\_Ethi → tha\_Thai} &2.22 / 0.57 &4.15 / 7.19 &6.59 / 8.71 &5.08 / 8.10 &2.29 / 3.96 &6.52 / 10.71 \\
\bottomrule
\end{xtabular}

%\caption{Generated by SpreadLaTeX}\label{tab:full_results}
\normalsize
\twocolumn

\subsection{Analysis} \label{sec:app_anal}

\begin{figure}[!htbp] 
    \centering
    \includegraphics[width=0.5\textwidth, height=0.25\textheight]{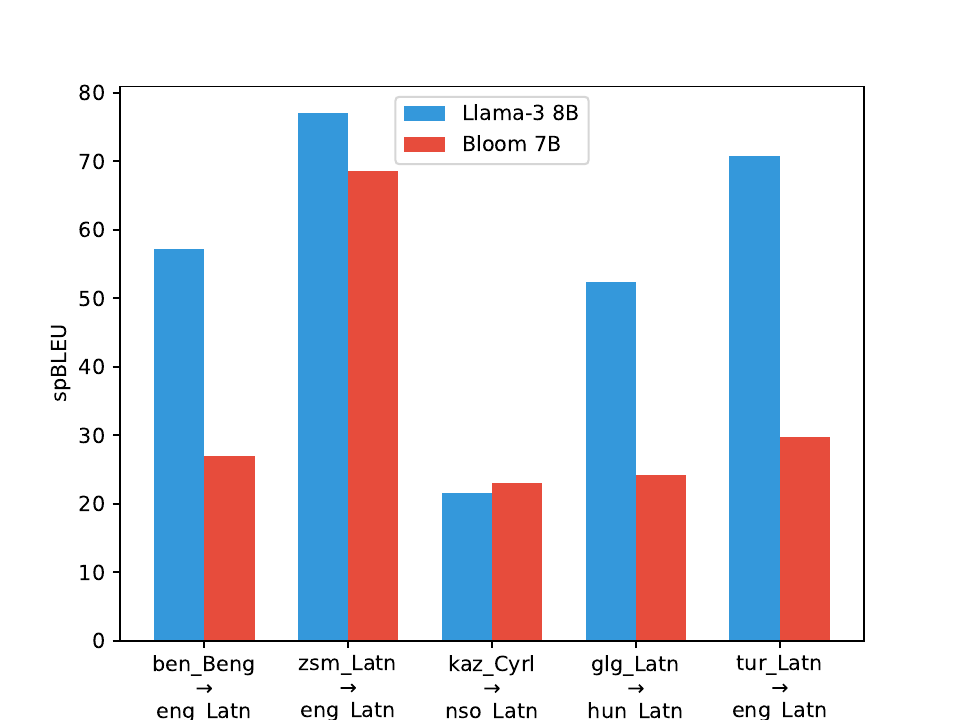} 
    \caption{Comparison of the spBLEU score of our unsupervised TopK+BM25 using Llama-3 and Bloom for a randomly selected subset of language pairs.}  %\textcolor{red}{(Full results in Table \ref{tab:iter_anal} in Appendix \ref{sec:app_anal}.)}} 
    \label{fig:spbleu_llms}
\end{figure}

\paragraph{Does the multilingual LLM testbed affect translation quality?} Our experiments were conducted using Llama-3 and Bloom. It is important to note that translation performance is highly correlated with the choice of the LLM. Overall, Llama-3 outperforms Bloom in terms of translation spBLEU score. As illustrated in Figure \ref{fig:spbleu_llms}, Llama-3 consistently outperforms Bloom across various language pairs, notably in \textit{tur\_Latn → eng\_Latn} and \textit{ben\_Beng → eng\_Latn}. However, there are exceptions where Bloom slightly surpasses Llama-3, such as in the \textit{kaz\_Latn → nso\_Latn} pair.

\begin{figure}[] 
    \centering
    \includegraphics[width=0.5\textwidth]{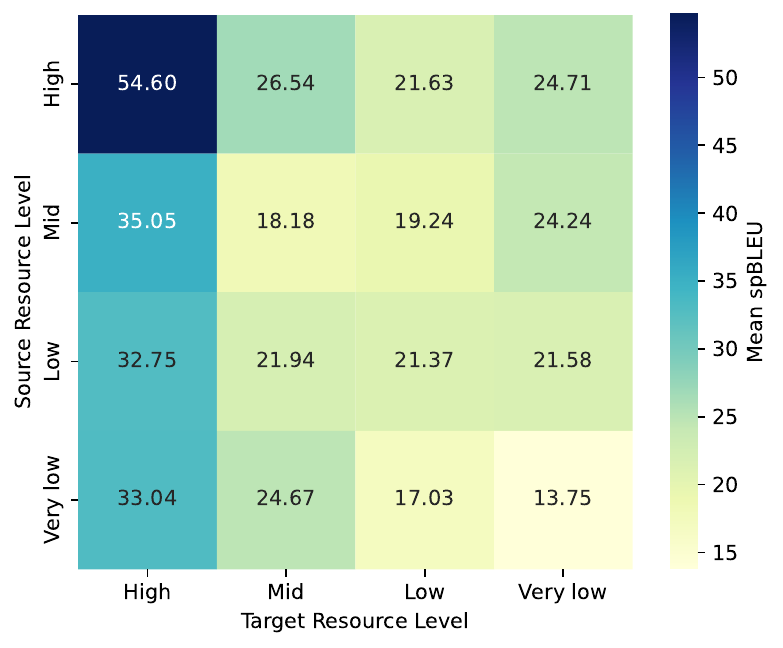} 
    \caption{Average spBLEU scores from TopK+BM25 experiments across different resource levels for different language pairs using Bloom. Each cell represents the mean spBLEU score for translations from a source resource level to a target resource level.} 
    \label{fig:spbleu_res_lev_agg_bloom}
\end{figure}

\begin{figure}[] 
    \centering
    \includegraphics[width=0.5\textwidth]{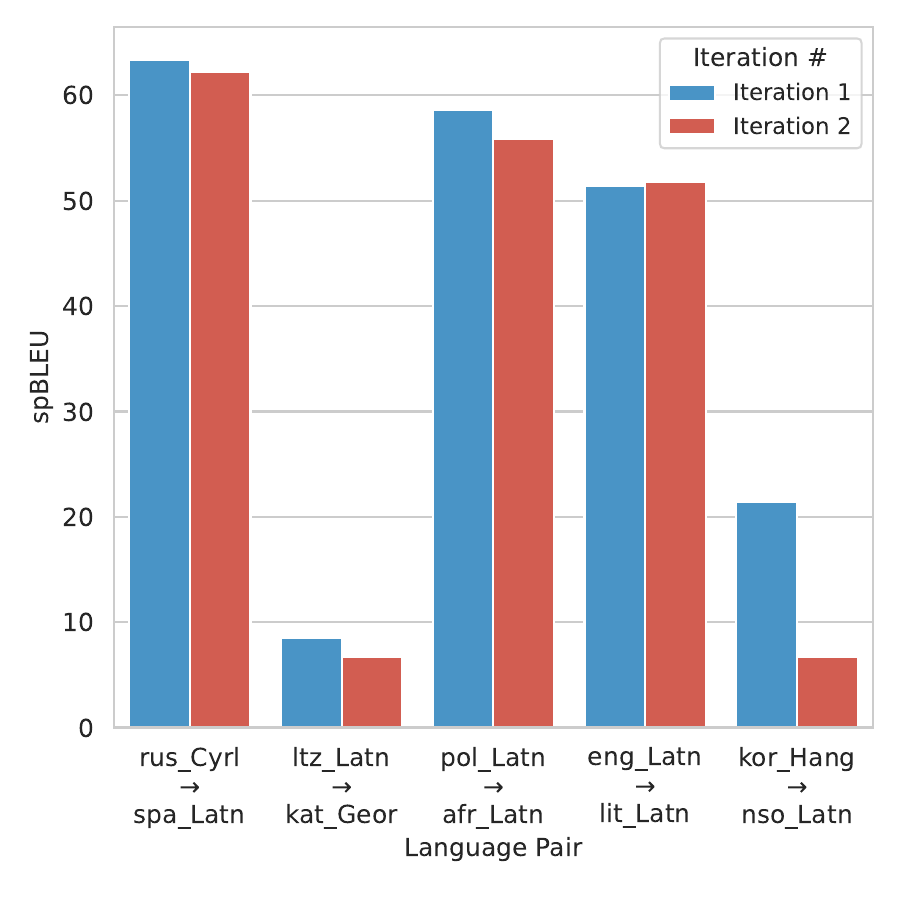} 
    \caption{The impact of multiple iterations of our approach using TopK+BM25 on the spBLEU score for a subset of language pairs using Llama-3.}  %\textcolor{red}{(Full results in Table \ref{tab:iter_anal} in Appendix \ref{sec:app_anal}.)}} 
    \label{fig:spbleu_its}
\end{figure}

\paragraph{How does iterative $k$-shot re-mining improve the quality of UMT?} Figure \ref{fig:spbleu_its} illustrates the spBLEU scores of five randomly selected language pairs from our experiments, where an additional iteration of the sentence-level translation phase described in Section \ref{sec:sent_trans} was performed. The results of this random subset indicate that, overall, the performance of UMT declines after another iteration of our algorithm. This can be observed notably in pairs such as \textit{kor\_Hang → nso\_Latn} and \textit{ltz\_Latn → kat\_Geor}, where the spBLEU performance using our UMT approach decreased considerably in the second iteration. Based on these findings, we conclude that applying our approach for a single iteration generally yields the best results.

\begin{figure}[!t] 
    \centering
    \includegraphics[width=0.5\textwidth]{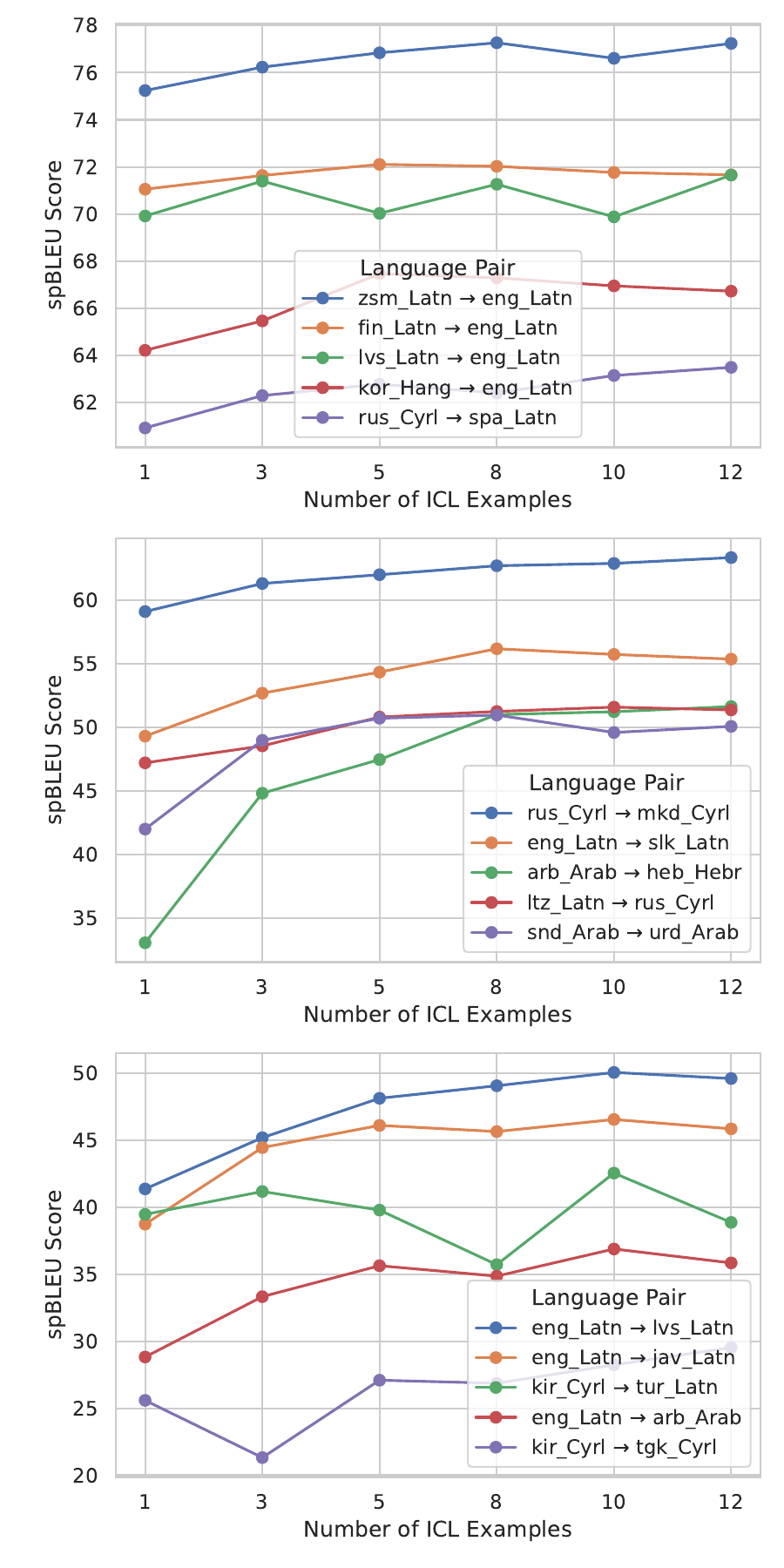} 
    \caption{The impact of the quantity of ICL examples on the spBLEU score for UMT employing our TopK+BM25 approach with Llama-3. } 
    \label{fig:spbleu_shots_full}
\end{figure}

\paragraph{Does the number of unsupervised ICL examples affect the UMT performance?} Figure \ref{fig:spbleu_shots_full} illustrates the influence of the number of chosen ICL examples on the spBLEU score when employing the TopK+BM25 selection approach. The figure indicates that selecting a greater number of ICL examples generally enhances performance. This effect is particularly significant between 1 and 8 examples, where the spBLEU score shows significant increases for the majority of pairs. However, the increments in spBLEU score become less substantial between 8 and 12 examples.

\iffalse
\begin{figure}[!htbp] 
    \centering
    \includegraphics[width=0.5\textwidth, height=0.25\textheight]{plots/spbleu_scores_k_shots.pdf} 
    \caption{The impact of the quantity of ICL examples on the spBLEU score for UMT employing our TopK+BM25 approach with Llama-3. (Results for additional language pairs are presented in Figure \ref{fig:spbleu_shots_full} in Appendix \ref{sec:app_anal}.)} 
    \label{fig:spbleu_shots}
\end{figure}
\fi
\paragraph{Why does the translation performance in TopK ICL decrease compared to Random ICL?} Table \ref{tab:main_results} illustrates a decline in performance when selecting the most confident pairs (\textit{TopK ICL)} compared to \textit{Random ICL}. Our analysis suggests that this decrease occurs because the similarity function prioritizes semantic confidence while disregarding linguistic details, consequently pairing examples from identical languages at times. In contrast, the BM25 selection method integrates both semantic relevance and surface-level linguistic comparisons. This approach results in pairs that not only represent the source and target languages accurately but also maintain semantic similarity.

\begin{figure}[!htbp] 
    \centering
    \includegraphics[width=0.5\textwidth, height=0.25\textheight]{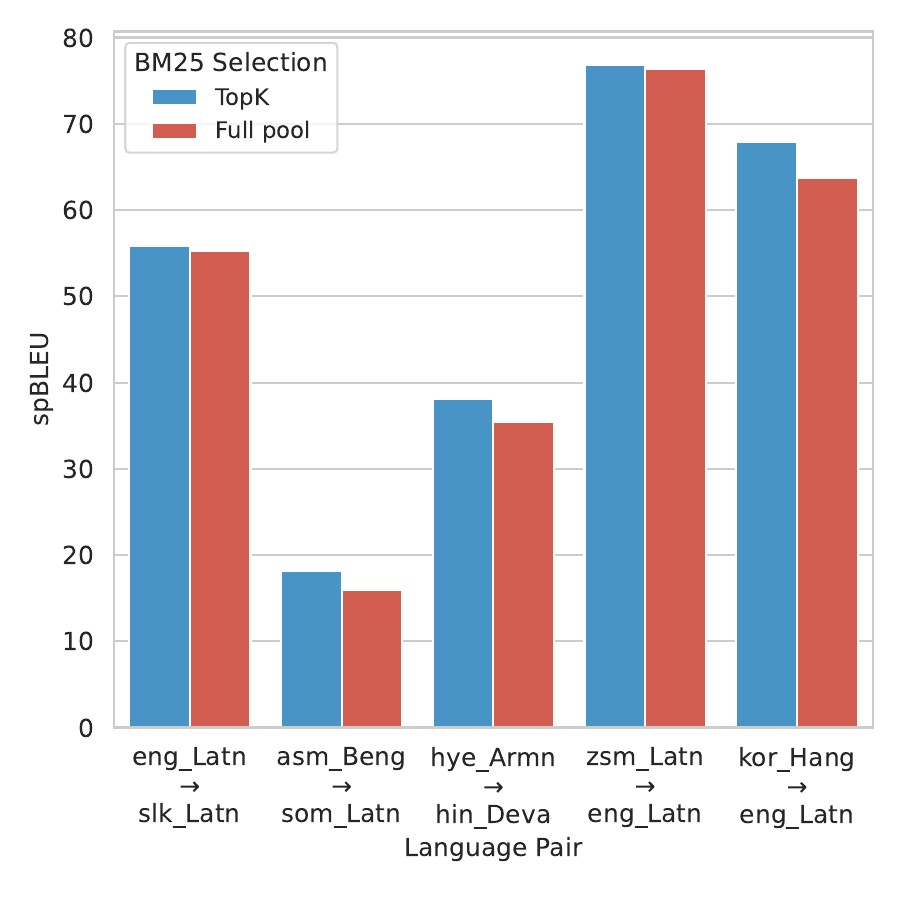} 
    \caption{The impact of selecting the unsupervised $k$-shot from the most confident pairs (TopK) compared to selecting from all pairs (Full pool) on the spBLEU score for a subset of language pairs using Llama-3. (Table \ref{tab:shots_bm25_anal} in Appendix \ref{sec:app_anal} presents results for additional pairs.)} 
    \label{fig:spbleu_bm25}
\end{figure}

\paragraph{Does applying BM25 filtering to the most similar sentence pairs enhance the quality of the resulting in-context examples?} In our TopK+BM25 results, we applied our BM25 filtering criteria exclusively to sentence pairs that exceed a specific similarity threshold, $\tau$. %This analysis investigates whether restricting the sentence pair pool enhances performance.
Figure \ref{fig:spbleu_bm25} compares the results between applying BM25 selection on the entire pool of mined sentences (full pool) and applying it only to pairs filtered by the similarity threshold (TopK). The results indicate that using BM25 on the most similar parallel pairs generally yields higher spBLEU scores. This improvement is particularly noticeable in pairs such as \textit{kor\_Hand → eng\_Latn} and \textit{asm\_Beng → som\_Latn}. Table \ref{tab:shots_bm25_anal} in Appendix \ref{sec:app_anal} presents results for 19 language pairs, demonstrating that the TopK+BM25 selection strategy outperforms the application of BM25 on the full pool, with average spBLEU scores of $48.80$ and $48.30$, respectively.

\paragraph{What is the impact of the selection threshold $\tau$ on the TopK+BM25 performance?} Our proposed TopK+BM25 in-context selection for UMT depends on the threshold $\tau$ to filter only the most confident semantically similar pairs. We conduct experiments varying the similarity threshold $\tau$ to analyze its impact on in-context example quality and translation performance. Our findings across multiple language pairs are presented in Table \ref{tab:tau-mt-scores}. From this subset, we conclude that the selection of the threshold $\tau$ has an impact on the TopK+BM25 spBLEU score, with higher $\tau$ values generally improving spBLEU scores, indicating better translation quality.

\begin{table}[t]
\centering

\resizebox{\columnwidth}{!}{%
\begin{tabular}{@{}l@{\hspace{4pt}}c@{\hspace{4pt}}c@{\hspace{4pt}}c@{\hspace{4pt}}c@{\hspace{4pt}}c@{}}
\toprule
\textbf{Lang. Pair} & \multicolumn{5}{c}{\textbf{Threshold $\tau$}} \\
 & \textbf{0.0} & \textbf{0.3} & \textbf{0.5} & \textbf{0.7} & \textbf{0.9} \\
\midrule
\textbf{ita\_Latn} & 67.89 & 67.93 & 68.03 & 68.13 & 68.34 \\
\textbf{cat\_Latn} & 71.92 & 72.14 & 72.76 & 72.81 & 73.95 \\
\textbf{zsm\_Latn} & 70.67 & 70.98 & 71.01 & 71.50 & 71.20 \\
\textbf{bel\_Cyrl} & 44.37 & 44.65 & 44.81 & 45.58 & 45.39 \\
\bottomrule
\end{tabular}%
}
\caption{Impact of Threshold $\tau$ on our TopK+BM25 UMT spBLEU Scores using Llama-3. These experiments were conducting by having the English language as source and the cited language in the Lang. Pair column as target.}
\label{tab:tau-mt-scores}
\end{table}

\begin{figure}[] 
    \centering
    \includegraphics[width=0.5\textwidth, height=0.25\textheight]{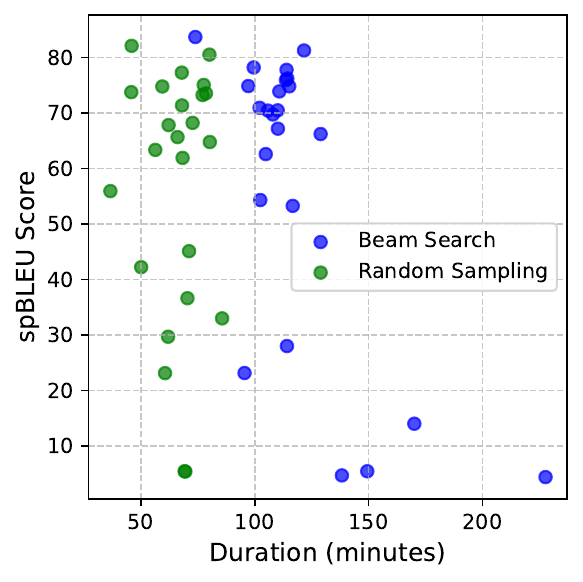} 
    \caption{spBLEU scores and durations (in minutes) for the MT performance achieved using TopK+BM25 with Llama-3.} 
    \label{fig:duration_decoding}
\end{figure}

\paragraph{Does the decoding strategy affect performance?} We performed the experiments in Table \ref{tab:main_results}, but using Beam Search instead of Random Sampling. Results are presented in Figure \ref{fig:duration_decoding} and compare both decoding strategies. Experiments performed using Beam Search are indeed slower than those using Random Sampling; however, the scores resulting from both models are comparable. This led us to choose random sampling in all our experiments.

\paragraph{Examples of translations made using our approach compared to the ground truth references and the regular $k$-shot baseline.} Table \ref{tab:predictions-examples} presents a sample of translations generated using our TopK+BM25 ICL approach compared to the regular $k$-shot and our unsupervised word-by-word approach. The examples indicate that our TopK+BM25 ICL approach can generate translations of comparable or superior quality to those produced by the regular ICL. Additionally, the examples demonstrate that while the word-by-word translation approach preserves the overall meaning, it often results in grammatical errors and incorrect word order. We discussed this issue in Section \ref{sec:tran_step_1}, where we explained the necessity for a back-translation step to mitigate these problems and produce more natural language translations from these word-by-word translations.

\begin{table*}[!htp]\centering
\scriptsize
\begin{tabular}{p{2.5cm}p{3cm}p{9cm}}\toprule
\multirow{5}{*}{\textbf{ast\_Latn → eng\_Latn}} &\textbf{Source} &La corteya tien alredor de 70 km d’espesor nel llau más cercanu y de 100 km nel más llonxanu. \\\addlinespace[1mm]
&\textbf{Reference} &The crust is about 70 km thick on the near side and 100 km thick on the far side. \\\addlinespace[2mm]
&\textbf{Supervised $k$-shot} &The crown has a thickness of about 70 km on the closest side and 100 km on the farthest side. \\\addlinespace[1mm]
&\textbf{Unsupervised W2W} &the cut ten around of 70 km d’espesor in cry more close and of 100 km in more llonxanu. \\\addlinespace[1mm]
&\textbf{TopK+BM25} &The crust has about 70 km of thickness on the near side and 100 km on the far side. \\\midrule
\multirow{5}{*}{\textbf{kir\_Cyrl → eng\_Latn}} &\textbf{Source} &Agencija Xinhua izvijestila je da su vladini istražitelji u srijedu pronašli dva snimača podataka o letu iz „crne kutije“. \\\addlinespace[1mm]
&\textbf{Reference} &Xinhua reported that government investigators recovered two 'black box' flight recorders on Wednesday. \\\addlinespace[2mm]
&\textbf{Supervised $k$-shot} &Chinese state investigators found two flight recorders in a "black box" on Wednesday, according to the Xinhua news agency. \\\addlinespace[1mm]
&\textbf{Unsupervised W2W} &agency xinhua reported yes yes you governmental researchers in wednesday find two cameras data o flight from „crne kutije“. \\\addlinespace[1mm]
&\textbf{TopK+BM25} &Agency Xinhua reported that government investigators had found two flight recorders from "the black box" on Wednesday. \\\midrule
\multirow{5}{*}{\textbf{cat\_Latn → eng\_Latn}} &\textbf{Source} &Vladajuća stranka, Narodna organizacija jugozapadne Afrike (South West Africa People's Organisation, SWAPO), također je zadržala većinu na parlamentarnim izborima. \\\addlinespace[1mm]
&\textbf{Reference} &The ruling party, South West Africa People's Organisation (SWAPO), also retained a majority in the parliamentary elections. \\\addlinespace[2mm]
&\textbf{Supervised $k$-shot} &The ruling party, the South West Africa People's Organisation (SWAPO), also kept its parliamentary majority. \\\addlinespace[1mm]
&\textbf{Unsupervised W2W} &ruling stranka, national organization southwestern africa (South west africa People's Organisation, SWAPO), also yes kept most on parliamentary izborima. \\\addlinespace[1mm]
&\textbf{TopK+BM25} &The ruling party, the South West Africa People's Organisation (SWAPO), also retained a majority in parliamentary elections. \\\midrule
\multirow{5}{*}{\textbf{hrv\_Latn → eng\_Latn}} &\textbf{Source} &Velika područja dalje na sjeveru vrlo su rijetko naseljena, a nekima gotovo vlada nenaseljena divljina. \\\addlinespace[1mm]
&\textbf{Reference} &Large areas further north are quite sparsely populated and some is nearly uninhabited wilderness. \\\addlinespace[2mm]
&\textbf{Supervised $k$-shot} &Large areas further north are sparsely inhabited, and some remain wild. \\\addlinespace[1mm]
&\textbf{Unsupervised W2W} &great areas further on north very you rarely naseljena, a some almost government uninhabited divljina. \\\addlinespace[1mm]
&\textbf{TopK+BM25} &Large areas further north are extremely sparsely populated and some are virtually wilderness. \\
\bottomrule
\end{tabular}
\caption{Examples of translations made by our proposed unsupervised Word-by-word (W2W) and TopK+BM25 ICL methods compared with regular $k$-shot ICL using Llama-3.}\label{tab:predictions-examples}

\end{table*}

\begin{table}[!htp]\centering
\scriptsize
\begin{tabular}{lrrr}\toprule
\textbf{Pair} &\textbf{TopK Pairs} &\textbf{Full pool} \\\midrule
\textbf{eng\_Latn → slk\_Latn} &\textbf{55.90} &55.30 \\
\textbf{kor\_Hang → eng\_Latn} &\textbf{67.90} &63.80 \\
\textbf{eng\_Latn → jav\_Latn} &46.50 &\textbf{49.20} \\
\textbf{cat\_Latn → ckb\_Arab} &19.70 &\textbf{23.60} \\
\textbf{eng\_Latn → lvs\_Latn} &49.90 &\textbf{50.60} \\
\textbf{fin\_Latn → eng\_Latn} &\textbf{71.30} &69.70 \\
\textbf{rus\_Cyrl → spa\_Latn} &\textbf{62.80} &62.00 \\
\textbf{rus\_Cyrl → mkd\_Cyrl} &63.10 &\textbf{63.40} \\
\textbf{arb\_Arab → heb\_Hebr} &\textbf{51.20} &47.10 \\
\textbf{lvs\_Latn → eng\_Latn} &\textbf{70.80} &68.70 \\
\textbf{kir\_Cyrl → tgk\_Cyrl} &\textbf{26.90} &23.80 \\
\textbf{kat\_Geor → hun\_Latn} &31.70 &\textbf{40.40} \\
\textbf{asm\_Beng → som\_Latn} &\textbf{18.20} &15.90 \\
\textbf{snd\_Arab → urd\_Arab} &50.30 &\textbf{50.40} \\
\textbf{ltz\_Latn → rus\_Cyrl} &\textbf{52.10} &48.40 \\
\textbf{hun\_Latn → kat\_Geor} &32.30 &\textbf{34.70} \\
\textbf{hye\_Armn → hin\_Deva} &\textbf{38.10} &35.40 \\
\textbf{kir\_Cyrl → tur\_Latn} &\textbf{40.70} &37.90 \\
\textbf{zsm\_Latn → eng\_Latn} &\textbf{76.90} &76.40 \\\midrule
\textbf{Average} &\textbf{48.80} &48.30 \\
\bottomrule
\end{tabular}
\caption{spBLEU scores evaluating the performance of our unsupervised sentence-level translation model using LLaMa-3. The scores compare two methods: selecting the top-k parallel sentences with the highest similarity scores between source and target languages using BM25 (TopK), and applying BM25 to the entire pool of parallel sentences (Full Pool).}\label{tab:shots_bm25_anal}
\end{table}

\twocolumn

\end{document}